# Automatic Detection of ECG Abnormalities by using an Ensemble of Deep Residual Networks with Attention

Yang Liu[1+], Runnan He[1+], Kuanquan Wang[1], Qince Li[1], Qiang Sun[5], Na Zhao[1] and Henggui Zhang[1,2,3,4*]

[1] School of Computer Science and Technology, Harbin Institute of Technology (HIT), 150001 Harbin, China
[2] School of Physics and Astronomy, The University of Manchester, Manchester M13 9PL, UK
[3] SPACEnter Space Science and Technology Institute, Shenzhen 518117, China
[4] International Laboratory for Smart Systems and Key Laboratory of Intelligent of Computing in Medical Image, Ministry of Education, Northeastern University, Shengyang, 110004, China
[5] The Department of Pharmacology, Beijing Electric Power Hospital, Beijing, 100073, China
[+]Joint first author
[*]henggui.zhang@manchester.ac.uk

**Abstract.** Heart disease is one of the most common diseases causing morbidity and mortality. Electrocardiogram (ECG) has been widely used for diagnosing heart diseases for its simplicity and non-invasive property. Automatic ECG analyzing technologies are expected to reduce human working load and increase diagnostic efficacy. However, there are still some challenges to be addressed for achieving this goal. In this study, we develop an algorithm to identify multiple abnormalities from 12-lead ECG recordings. In the algorithm pipeline, several preprocessing methods are firstly applied on the ECG data for denoising, augmentation and balancing recording numbers of variant classes. In consideration of efficiency and consistency of data length, the recordings are padded or truncated into a medium length, where the padding/truncating time windows are selected randomly to suppress overfitting. Then, the ECGs are used to train deep neural network (DNN) models with a novel structure that combines a deep residual network with an attention mechanism. Finally, an ensemble model is built based on these trained models to make predictions on the test data set. Our method is evaluated based on the test set of the First China ECG Intelligent Competition dataset by using the $F_1$ metric that is regarded as the harmonic mean between the precision and recall. The resultant overall $F_1$ score of the algorithm is 0.875, showing a promising performance and potential for practical use.

## 1 Introduction

Heart diseases, mainly manifested as disordered patterns of atrial and ventricular electrical excitation activity, have been regarded as the leading cause of morbidity and mortality. Electrocardiogram (ECG) is a common and noninvasive tool that can be used for diagnosing heart conditions. However, it is time consuming and error-prone to analyze ECGs in practice, therefore, computer-aided algorithms may offer a promising way to improve the efficiency and accuracy of ECG analyzing.

The algorithms for ECG analyzing typically contain three steps, which are preprocessing, feature extraction and classification [1-5]. Among these, the feature extraction is a critical step, for which many methods have been proposed, such as morphology information [2], temporal and frequency features [3], high order statistical features [4] and wavelet features [5]. However, these algorithms still have shortcomings to achieve a good performance for the detection of abnormalities in ECGs. Recently, deep convolutional neural networks (DNNs) showed outstanding performance in automatic feature extraction, leading to a dramatic breakthrough in a range of fields associated with computer vision [6]. Therefore, in the field of ECG analysis, many studies have attempted to apply DNNs, such as convolutional neural network (CNN) [7], deep residual network [6], and recurrent neural network (RNN) [8], to address the problem of heart diseases detection, all of which have achieved some impressive results. However, due to the long recording length, low signal quality and pathological diversity of ECG recordings, it is still a challenge for accurate feature extraction. .

In this study, we propose a novel deep residual neural network with attention mechanism to detect a series of abnormalities from 12-lead ECG recordings. The deep residual neural network is used to learn local features from the ECG waveforms, while the attention mechanism determines the relevance of features from each part and summarizes them into a single feature vector that is used for the classification. This combination demonstrates to be effective for feature learning in a long ECG recording, and robust when the signal is partially corrupted.

## 2  Materials

The First China ECG Intelligent Competition dataset contains about 15000 12-lead ECG recordings, among which 6,500 for training and 8,500 for testing respectively. The recordings are in different lengths, ranging from 9 to 90 seconds sampled at 500 Hz (Fs=500Hz). Each recording is labeled with one or more types including normal, atrial fibrillation (AF), first degree atrioventricular block (FDAVB), complete right bundle branch block (CRBBB), left anterior fascicular block (LAFB), premature ventricular contraction (PVC), premature atrial contraction (PAC), early repolarization pattern changes (ER) and T-wave changes (TWC).

## 3  Method

### 3.1  Preprocessing

**Baseline Wander Removal.** Baseline wander results from low-frequency noise in the ECG signal. It can influence the diagnosis of many diseases that manifest as low-frequency changes in the ECG signals, e.g., S-T segment changes. We remove the baseline wander by first estimating it and then subtracting it from the original signal. The estimating is based on moving average which is a windowed low-pass filter with the cut-off frequency calculated by

$$f_{co} = 0.443 \times \frac{f_s}{N} \tag{1}$$

where $f_{co}$ indicates the cut-off frequency, $f_s$ indicates the sampling frequency, and N indicates the window size. Generally, the cut-off frequency shouldn't be less than the slowest heart rate which is typically 40 beats/minute, i.e., 0.67Hz. But, considering the fluctuation of heart rate, the cut-off frequency should be slightly lower, approximately 0.5Hz.

**Powerline Interference and Muscle Noise Removal.** Powerline interference generally resulted from the alternating current (AC) in the environment. Thus, its frequency is usually 50/60Hz depending on the specific standard of the AC power supply system. Muscle noise, i.e., electromyographic noise, is caused by the electrical activity produced by muscle contraction. Different from the powerline interference, muscle noise is much more irregular, due to the randomness of muscle activities. The frequency components of muscle noise have a wide overlap with those of the ECG, and can be even higher. In this work, we remove both these noises by wavelet denoising based on 5-level 'db4' wavelet transform and soft-thresholding [9].

**Padding or Truncating Signals to the Same Length.** The lengths of ECG recordings in the dataset is in a high variety, ranging from 9s to 90s. For batch processing of a DNN model, the recordings in a batch should be in the same length. To address this problem, there are three ways:

- Padding all recordings into the longest length. This method avoids the loss of information during the length-unifying process. But, for most of the recordings, the new length is several times longer than their original lengths, which will result in the significant increase in processing time of a DNN model.
- Truncating signals into the shortest length. On the contrary to the padding method, this method can reduce the processing time significantly. However, the truncating process will inevitably lead to the loss of information, especially when the truncated part is in a large proportion of the original signal.
- Grouping the recordings that have the same length. Compared to the above two methods, this method doesn't change the original signals and thus won't cause any loss or distortion to their information. But, as the distribution of the recording lengths is extremely uneven, a group may contain just one or two recordings, resulting in a big variety of batch sizes. Besides, there is some uncertainty in the prediction by a DNN model when it receives a recording with an unknown length.

In this work, we try to make a trade-off between the recording integrity and computing complexity by padding or truncating the signals into a medium length. As more than 90% of recordings in the dataset is no longer than 30s, we choose 30s (i.e., 15000 sampling points) as the target length. To minimize the effect of padding on the following interpretations, the value for padding is set to zero which is equal to the baseline of the ECG signals. The obvious difference between the padded part and the original part will help a machine learning model to detect the original part from the whole recording. However, the padding and truncating operations can be done in different positions, which may lead to different impacts on the interpretation by a DNN model. We will discuss this in the following section about data augmentation and balancing.

**Redistribution of Signal Lengths.** Even though the recordings are padded or truncated into the same length, the difference of original length distributions between data classes can still induce a bias to the discrimination of a DNN model. For example, the recordings longer than 20s only account for a proportion less than 1% in the normal class, but account for more than 10% in the PAC class. As a result, a DNN model may recognize the padding length as a feature to distinguish between these classes, which is clearly unreasonable. Therefore, we propose a method to organize the recordings into the same distribution of recording lengths among all the classes. We first make a distribution statistics of recording lengths in the whole dataset. Because of the truncating operation in our pipeline as stated above, recordings longer than 30s are all counted as that of 30s. Then, the global distribution is used as the target distribution, and recordings in each class are augmented to have the same distribution. For lengths that exist in the target distribution but not exist in the original distribution of a class, we truncate the longer recordings to make recordings with these lengths.

**Data Augmentation and Balancing.** As discussed above, there are different ways for padding or truncating a recording in terms of time windows. In order to make a DNN model insensitive to the timing positions of padding or truncating, we pad or truncate each recording at different positions to create more samples for training. In other words, we augment the dataset by different padding and truncating ways. The selection for padding or truncating way is random in our study. For a recording shorter than the target length, we can pad it at both the ends with various of schemes to specify the padding length at each end. This method introduces more randomness to the positions of padding, which would help a DNN model learn to ignore the padded parts and focus the original parts of the recordings. And for the truncated recordings, this method generates more training samples with different parts of the original recordings that would contribute to better use of the limited data and enhance the model's discrimination ability. Besides, in terms of recordings' numbers, the dataset is very imbalanced between classes. The data augmentation method can also be used to balance the dataset. Generally speaking, recordings with short class length will be augmented more times than those with long class length, allowing each class have approximately the same number of recordings.

## 3.2 Model Architecture

Due to the automatic feature-learning ability, DNNs can reduce human working load in extracting features from the raw ECGs. A DNN model is supposed to learn a brief, robust but comprehensive representation from a raw ECG recording. In this work, we propose a novel DNN architecture that combines a residual convolutional network and an attention mechanism, as shown in Fig. 1.

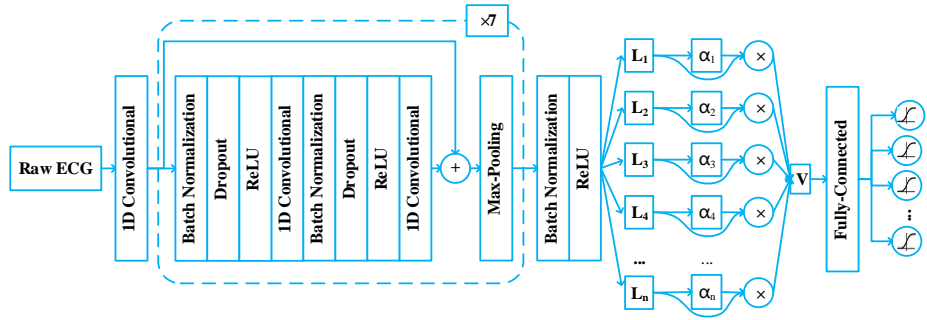

**Fig. 1.** The architecture of proposed deep residual network with attention mechanism. $L_i$ indicates the i-th local feature vector. $\alpha_i$ indicates the attention value for the i-th local feature vector. V indicates the global feature vector.

In our proposed architecture, the feature learning process can be divided into two stages: local features learning stage and global features learning stage. A local feature vector learned by a stack of residual convolutional modules characterizes a short fragment of a ECG recording, while the global feature vector learned by an attention mechanism is a summary of the sequence of local-feature vectors. After the feature learning, the global feature vector is input into a fully-connected layer to predict the probabilities that a recording belongs to each class. We will give more details about this architecture in the following section.

In the local features learning stage, a raw ECG recording is first input into a 1D convolutional layer. The output feature map is then processed sequentially by 7 residual convolutional modules which are considered having good properties to avoid the degradation problem in DNNs [6]. Each residual model is constructed by 9 layers: 2 batch normalization layers, 2 dropout layers, 2 ReLU activation layers, 2 1D-convolutional layers and an addition-based merging layer, in the order shown in Fig.1. The kernel size of each convolutional layer is 16. The kernel number in the first convolutional layer is 16, and it grows by 16 for every two residual modules. There is also a max-pooling layer following each residual module for compression of intermediate feature maps. As a result, the length of a feature map output by the local feature learning part will be $1/2^7$ of the input length.

In the global features learning stage, an attention mechanism is utilized to learn an attention distribution on the sequence of local features. Due to the possible paroxysm of diseases, padding parts and noise effects, only a few episodes in a ECG recording may be relevant for the diagnosis. In view of this, the attention distribution is supposed to manifest the relevance of each part in the ECG recordings for the classification. Then, the local features are summed, weighted by the attention, into a single feature vector. Finally, a fully-connected layer is used to learn a classifier based on the global features. This layer contains 9 cells corresponding to the 9 categories respectively. As a record may belong to more than one category, the output of each cell is processed by a sigmoid activation function to make prediction independently.

### 3.3 Model Training

Based on the architecture stated above, we train a series of models with different procedures. As shown in Fig. 2, there are 4 different pipelines (labeled with numbers) are used in our research for model training. Most of the differences between pipelines are present in the preprocessing steps, including data normalization, denoising, data augmentation and balancing between classes. The window size for baseline wander removal is 250 ($f_{co} = 0.886$Hz) in pipeline 1, while it is 500 ($f_{co} = 0.443$Hz) in other pipelines for denoising. In the pipeline 2, recordings from the CPSC 2018 dataset are mixed into the training dataset of this challenge to enhance the robustness and generalization ability of the trained models. Besides, as the recordings of training set and test set are in the same distribution which is imbalanced among data classes, the balancing operations to the training set may cause a bias to the trained models and hence reduce the classification accuracy. Therefore, we utilize a two-stage (balance and imbalanced) strategy to address this problem. Specifically, in pipelines 3, the models trained with balanced data (from pipeline 2) are retrained with imbalanced data which is in the same distribution with the original training set. The balanced data helps the model to learn more distinguishing features, while the data in the independent identically distribution with the test set helps to improve the classification accuracy. In pipeline 4, the input data is not denoised but normalized to have zero mean and unit variance. The model training steps in these pipelines are all in a 10-fold cross-validation manner. Furthermore, the training/testing samples division in each fold is also in the same scheme among these pipelines, which can especially avoid overlap between training and testing samples in the two-stage model training. At the final step, an ensemble model is built by averaging the probabilistic predictions of models from all these pipelines.

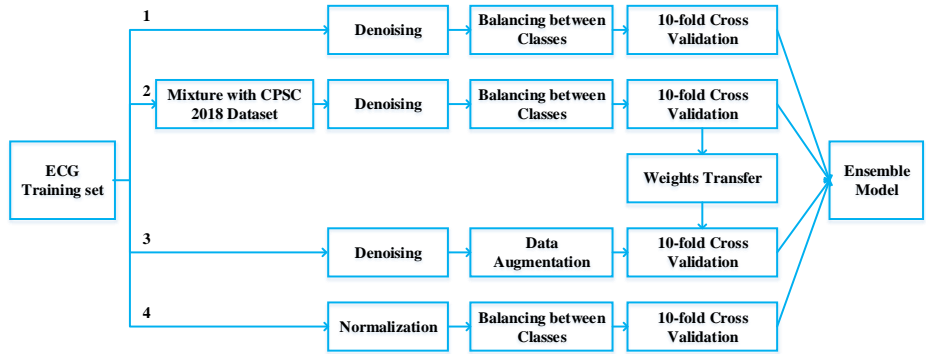

**Fig. 2.** The model training workflow that combines models from different pipelines to make an ensemble model.

## 4 Results and Discussion

The metrics are designed based on the evaluation of multi-label classification, where a single recording may belong to more than one class. The predictive accuracy of the algorithm for each class is measured by the $F_1$ score. Besides, an overall $F_1$ score is also calculated by averaging all the categorical sub-scores. The formulas for calculation of these metrics are described in the following.

For each class ($0 \leq j \leq 8$), there are four values counting samples with different prediction results, namely true positive (TP), false positive (FP), true negative (TN) and false negative (FN):

$$TP_j = |\{x_i | y_j \in Y_i, y_j \in f(x_i), 1 \leq i \leq N\}|, \tag{2}$$

$$FP_j = |\{x_i | y_j \notin Y_i, y_j \in f(x_i), 1 \leq i \leq N\}|, \tag{3}$$

$$TN_j = |\{x_i | y_j \notin Y_i, y_j \notin f(x_i), 1 \leq i \leq N\}|, \tag{4}$$

$$FN_j = |\{x_i | y_j \in Y_i, y_j \notin f(x_i), 1 \leq i \leq N\}|. \tag{5}$$

where $x_i$ indicates a sample for prediction, $y_j$ is the label for the class j, $Y_i$ is the annotated label set of $x_i$, and $f(x_i)$ is the predicted label set of $x_i$. The precision, recall and $F_1$ of each class can be calculated by

$$Precision_j = \frac{TP_j}{TP_j + FP_j}, \tag{6}$$

$$Recall_j = \frac{TP_j}{TP_j + FN_j}, \tag{7}$$

$$F_{1j} = \frac{2 \cdot Precision_j \cdot Recall_j}{Precision_j + Recall_j}. \tag{8}$$

The overall $F_1$ score is the arithmetic mean value of that of the nine classes.

$$F_1 = \frac{1}{9} \sum F_{1j} \tag{9}$$

The results show that the overall $F_1$ score of the proposed classifier on the hidden test set is 0.875, with the detailed scores shown in Table 1.

**Table 1.** Results of the ECG abnormalities classification on the entire test set.

|       | Normal | AF    | FDAVB | CRBBB | LAFB  | PVC   | PAC   | ER    | TWC   | Total |
|-------|--------|-------|-------|-------|-------|-------|-------|-------|-------|-------|
| $F_1$ | 0.875  | 0.974 | 0.901 | 0.983 | 0.747 | 0.971 | 0.926 | 0.736 | 0.757 | 0.875 |

Results shown in Table 1 demonstrate that the classifier achieves a good performance for AF, FDAVB, CRBBB, PVC and PAC, which are all above 0.9. However, the identification of LAFB, ER and TWC are less good, which are just over 0.7 due to relatively few data.

## 5     Conclusions

In this paper, two contributions have been made for ECG automatic classifications. (1) The random padding/truncating method is a simple strategy that not only helps to balance the processing efficiency and recordings integrity, but also provides ways to augment and balance the dataset. Furthermore, the randomness involved by this method in the padding/truncating positions can prevent a DNN overfitting the padding/truncating manner. (2) The proposed workflow that combines different pipelines of model training to make an ensemble model achieved better results than each of the separated pipelines. Classification results showed that the proposed algorithm may provide a potential way of computer-aided diagnosis for clinical applications.

### Acknowledgements

The work is supported by the National Science Foundation of China (NSFC) under Grant Nos. 61572152 (to HZ), 61571165 (to KW), 61601143 (to QL) and 81770328 (to QL), the Science Technology and Innovation Commission of Shenzhen Municipality under Grant Nos. JSGG20160229125049615 and JCYJ20151029173639477 (to HZ), and China Postdoctoral Science Foundation under Grant Nos. 2015M581448 (to QL).